\documentclass{article}
\usepackage{spconf,amsmath,graphicx}
\usepackage{times}
\usepackage{epsfig}
\usepackage{graphicx}
\usepackage{amssymb}
\usepackage{booktabs}
\usepackage{multirow}
\usepackage{makecell}
\usepackage{graphicx}
\usepackage{amssymb}
\usepackage{wrapfig}
 \usepackage{graphicx}
\usepackage{booktabs}
\usepackage{pgfplots} 
\usepackage{tikz}
\usepackage{subfig}
\usepackage{float}
\usepackage{color}
\usepackage{hyperref}
\hypersetup{
    colorlinks=true,
    linkcolor=blue,
    filecolor=blue,      
    urlcolor=blue,
    citecolor=cyan,
}

\title{CD-FSOD: A Benchmark for Cross-domain Few-shot Object Detection}
\name{Wuti Xiong}
\address{
  Center for Machine Vision and Signal Analysis, University of Oulu, Finland\\
  \texttt{wuti.xiong@oulu.fi}
}
\pgfplotsset{compat=1.17}
\makeatletter

\begin{document}
%
\maketitle
\begin{abstract}
 In this paper, we propose a study of the cross-domain few-shot object detection (CD-FSOD) benchmark, consisting of image data from a diverse data domain. On the proposed benchmark, we evaluate state-of-art FSOD approaches, including meta-learning FSOD approaches and fine-tuning FSOD approaches. The results show that these methods tend to fall, and even underperform the naive fine-tuning model.  We analyze the reasons for their failure and introduce a strong baseline that uses a mutually-beneficial manner to alleviate the overfitting problem. Our approach is remarkably superior to existing approaches by significant margins (2.0\% on average) on the proposed benchmark. Our code is available at \url{https://github.com/FSOD/CD-FSOD}.
\end{abstract}


\section{Introduction}
\label{sec:intro}

Few-shot object detection (FSOD) aims to detect novel classes of objects with a few annotated instances. In the previous FSOD setting \cite{kang2019few,fan2020few}, a detector is pre-training on the source dataset consisting of base classes and then transferred into the target dataset consisting of novel classes with few instances, where base classes and novel classes are disjoint but share similar data domains. However, this underlying assumption does not apply to some real-world scenarios because it is difficult or impossible to collect a sufficient amount of data in these domains. This leads to a new FSOD problem, where the detector must resort to pre-training in the base classes from a different domain. In these cases, even humans have trouble recognizing new categories that vary too greatly between examples or differ from prior experience \cite{lake2011one,lake2015human}. Thus, finding new approaches to tackle the problem remains a challenging but desirable goal.
\begin{figure}[!htp]
\centering
 \includegraphics[width= 0.98 \linewidth]{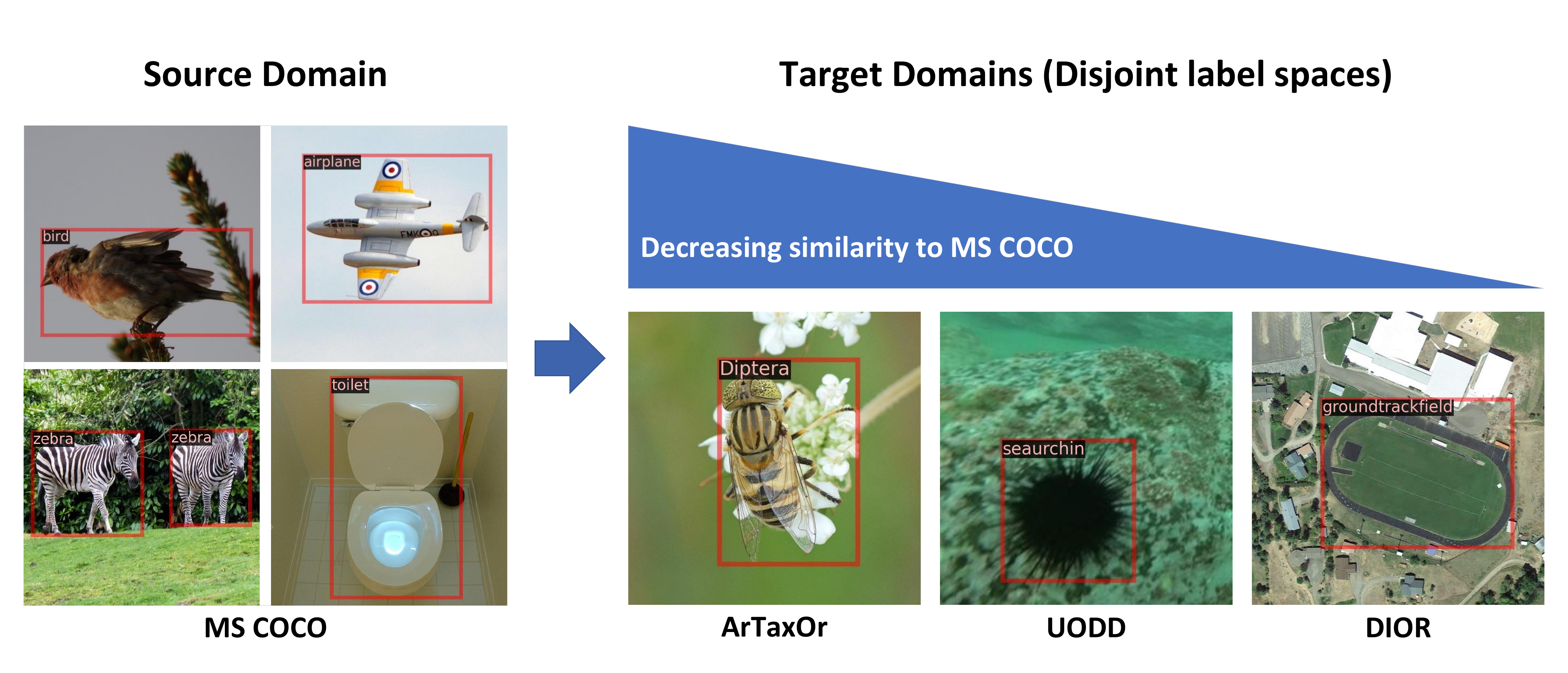}
 \vspace{-5pt}
\caption{The CD-FSOD benchmark. MS COCO \cite{lin2014microsoft} is used for source training, and domains of varying dissimilarity from natural images are used for target evaluation. }
\vspace{-5pt}
\label{fig:1}
\end{figure}

Although conventional FSOD benchmarks \cite{kang2019few,fan2020few} are well established, no works study FSOD across different domains. To fill this gap, In this paper, we introduce the study of \emph{\textbf{C}ross-\textbf{D}omain \textbf{F}ew-\textbf{S}hot \textbf{O}bject \textbf{D}etection} (CD-FSOD) benchmark (As shown in Figure \ref{fig:1}), which covers three target datasets: ArTaxOr \cite{arthropod}, UODD \cite{jiang2021underwater} and DIOR \cite{li2020object}. On the proposed benchmark, we conduct extensive experiments to evaluate existing FSOD approaches (including meta-learning approaches \cite{fan2020few,han2022meta,han2021query} and fine-tuning approaches \cite{wang2020frustratingly,DBLP:conf/cvpr/SunLCYZ21,qiao2021defrcn}). The results show that existing FSOD approaches can not achieve satisfactory performance and even underperform the naive fine-tuning model due to freezing parameters. Even without freezing parameters, fine-tuning methods struggle to outperform the naive transfer model while meta-learning methods still fail. This finding shows that existing FSOD methods cannot work for CD-FSOD, and there is an urgent need to develop new methods.

Besides, we introduce a novel distillation-based baseline, which enable a “flywheel effect” that the student and teacher can mutually reinforce each other so that both get better and better as the training goes on. Specifically, EMA (Exponential Moving Average) enables the teacher model to ensemble the student models in different time steps. The student’s weights are optimized by the distillation loss between the pseudo-labels generated by the teacher and the predictions by the students on the same image. Our approach outperforms existing FSOD approaches by a large margin on the proposed benchmark. In summary, our main contributions are as follows: (1)we established the CD-FSOD benchmark, where there is a very large domain difference between the base and target datasets; 2) on the proposed benchmark, we evaluate existing FSOD approaches, and analyze the reasons for their failure; 3) we introduce a strong baseline that achieves state-of-the-art performance on the proposed benchmark.

\section{Proposed Benchmark.}
\noindent \textbf{Rethinking FSOD Benchmarks.} 
In previous FSOD work \cite{kang2019few,fan2020few,han2021query,han2022meta,wang2020frustratingly,DBLP:conf/cvpr/SunLCYZ21,qiao2021defrcn}, two benchmarks have been widely adopted: MS COCO \cite{lin2014microsoft} and PASCAL VOC \cite{everingham2010pascal}. As for PASCAL VOC, there are three random split groups, and each of them covers 20 categories, which are randomly divided into 15 base classes and 5 novel classes. Each novel category has $K = 1, 2, 3, 5, 10$ objects sampled from the combination of VOC07 and VOC12 train/val set for few-shot detection training. As for MS COCO, the 60 categories disjoint with VOC are denoted as base classes while the remaining 20 classes are used as novel classes with $K = 1, 2, 3, 5, 10, 30$ shots. 5k images from the validation set are used for evaluation and the rest are used for training. While these benchmarks contributed to the research progress in FSOD, they have a limitation. As we discussed in Section \ref{sec:intro}, these benchmarks sample base classes and novel classes from a single dataset. There is a major issue that occurs commonly in practice: by the nature of the problem, collecting data from the same domain for many FSOD tasks is difficult. Under these circumstances, useful knowledge may still be effectively transferred across different domains, implying that approaches designed in the FSOD setting may not continue to perform well when applied to different domains, such as biological natural images and satellite images. Currently, no works study this scenario.

\noindent \textbf {CD-FSOD Benchmark.} To explore FSOD across a wide range of domains, we propose to build the benchmark using datasets from a wide range of domains rather than just a subset of natural image datasets. Our proposed benchmark include a base dataset (MS COCO \cite{lin2014microsoft} and 3 target datasets from diverse domains: ArTaxOr \cite{arthropod}, UODD \cite{jiang2021underwater} and DIOR \cite{li2020object}. The selected datasets reflect well-curated real-world use cases for few-shot object detection. In addition, collecting enough examples from the above domains is often difficult, expensive, or in some cases not possible. The similarity of these datasets to the MS COCO dataset, from high to low, is as follows: 1) ArTaxOr images are natural but are fine-grained (specific to biology); 2) UODD images are less similar as the poor visibility and low color contrast, but are still color images of natural scenes; 3) DIOR images are the most dissimilar as they have lost perspective distortion. The statistics of the target dataset are shown in Table \ref{tab:1}. Similar to the previous FSOD setting \cite{kang2019few,fan2020few}, the model is trained from a base dataset where each class has abundantly annotated instances, then is adapted to the target dataset where each class only has $K$ ($K = 1, 5, 10$) instances. 

\begin{table}[!htb]
\begin{center}
\resizebox{\linewidth}{!}{
  \begin{tabular}{cccccc}
    \toprule
     Domain &Dataset &  Classs &Train images&  Test images \\
         \midrule
    Biology&ArTaxOr   & 7  &13,991  &1,383 \\  
    Underwater &UODD      & 3  &3,194   &506\\
    Aerial&   DIOR      & 20 &18,463  &5,000\\
    \bottomrule
  \end{tabular} 
  }
  \end{center}
  \vspace{-10pt}
    \caption{The statistics of the target datasets. }
      \label{tab:1}
\end{table}

\section{Proposed Method} 
\begin{figure*}[!htp]
\centering
 \includegraphics[width= 0.9 \linewidth]{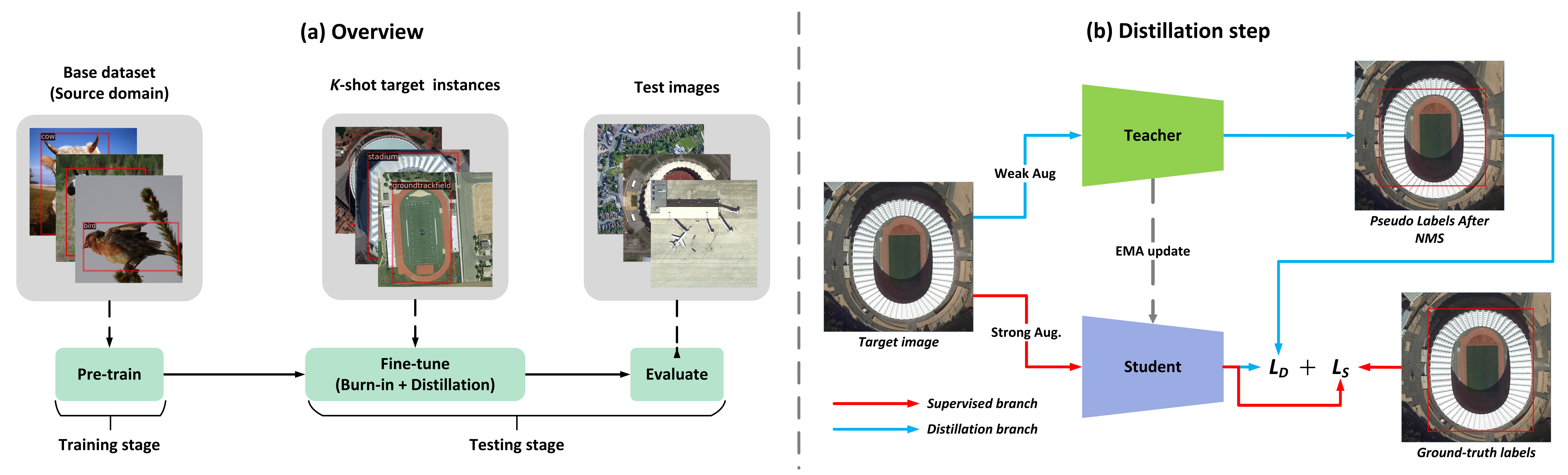}
 \vspace{-10pt}
\caption{(a) The overview of our proposed approach.  (b) Distillation step.}
\vspace{-10pt}
\label{fig:2}
\end{figure*}

\noindent \textbf {Overview.} As shown in Figure \ref{fig:2} (a), our approach consists of two stages: the training stage and the testing stage. At the training stage, we simply train the detector using the base data. At the beginning of the fine-tuning, we duplicate the initialized detector into the student model. The student first goes through a burn-in step, i.e. training the student with the standard detection supervision losses \cite{ren2016faster} on $K$-shot target instances. Then its weights are copied into the student and the teacher to initiate the distillation step. As shown in Figure \ref{fig:2} (b), in the distillation step, the teacher and student are trained in a mutually-beneficial manner, where the teacher promotes the student by the distillation loss, and its weights are updated by the student model via exponential mean average (EMA). The proposed method consists of two branches: the supervised branch and the distillation branch. The final loss $L$ is the sum of supervised loss $L_{S}$ and distillation loss $L_{D}$.
\begin{align}
  L & = L_{S} + \lambda L_{D} 
\end{align}
where the $\lambda $ is a hyper-parameter. As pointed out by prior works \cite{DBLP:conf/iclr/LiuMHKCZWKV21}, a key factor in improving the teacher is the diversity of student; thus, we use strongly augmented images as input for the student, but we use weakly augmented images as input to the teacher to provide reliable pseudo-labels.

\noindent \textbf {Supervised branch.} 
In the supervised branch, we compute the supervised detection losses (classification loss $ L_{cls}$
and localization loss $ L_{loc}$)  for the student model. With the $K$-shot target data $D_{s} = \left \{ x_{i}^{s}, y_{i}^{s}\right \}_{i=1}^{N_{s}}$, the supervised detection loss $L_{S}$ is written as:
\begin{align}
    L_{S} & =\sum_{i} L_{cls} (x_{i}^{s}, y_{i}^{s}) + L_{loc}(x_{i}^{s}, y_{i}^{s})
\end{align}

\noindent \textbf {Distillation branch.} As shown in Figure \ref{fig:2} (b), the teacher and student share the same architecture and are initialized with the same weights after the burn-in step. An image is processed independently by both the student and the teacher. The teacher is used to generate thousands of box candidates for the weakly augmented version. After NMS \cite{ren2016faster}) is  performed, only candidates with the foreground score higher than a threshold $\delta $ are retained as the pseudo boxes $p_{i}^{d}$. Then the distillation loss is obtained by calculating the detection loss between the student predictions $x_{i}^{d}$ and the pseudo-labels.

\begin{align}
   L_{D}  & = \sum_{i} L_{cls}(x_{i}^{d}, p_{i}^{d}) + L_{loc}(x_{i}^{d}, p_{i}^{d})
\end{align}

\noindent \textbf {Model update.} EMA update has been shown to be successful in many prior works \cite{kingma2015adam,ioffe2015batch,he2020momentum}. Thus, we use it to alleviate the overfitting problem in the CD-FSOD setting. Specifically, we detach the student and the teacher.  After obtaining the pseudo-labels from the teacher, only the learnable weights of the student $W_{s}$ is updated via back-propagation
\begin{align}
  W_{s} \leftarrow W_{s} +\gamma \frac{\partial L }{\partial W_{s} }
\end{align}
where the $\gamma$ denotes the learning rate. Then, the teacher model weights $W_t$ are updated from the student model weights $W_s$ by exponential moving average (EMA) \cite{tarvainen2017mean}. At each iteration, we update the teacher weights by:
\begin{align}
  W_{t} \leftarrow \alpha W_{t} + (1 - \alpha) W_{s}
\end{align}
where the $\alpha$ is a hyper-parameter.

\section{Experiment} 
\subsection{Implementation details} 
For a fair comparison, we follow previous work \cite{fan2020few,han2021query,han2022meta,wang2020frustratingly,DBLP:conf/cvpr/SunLCYZ21,qiao2021defrcn} to use Faster-RCNN \cite{ren2016faster} with FPN \cite{lin2017feature} and ResNet-50 backbone \cite{DBLP:conf/cvpr/HeZRS16} to build the student and teacher. For generating the pseudo boxes, we use confidence threshold $\delta $ = 0.7. For the data augmentation, we apply random horizontal flips for weak augmentation and randomly add color jittering, grayscale, Gaussian blur, and cutout patches for strong augmentations. Our implementation builds upon the Detectron2 framework. For the baselines\cite{fan2020few,han2021query,han2022meta,wang2020frustratingly,DBLP:conf/cvpr/SunLCYZ21,qiao2021defrcn}, we use the official implementations: A-RPN\footnote{\url{https://github.com/fanq15/FewX}},H-GCN\footnote{\url{https://github.com/GuangxingHan/QA-FewDet}},  Meta-RCNN\footnote{\url{https://github.com/guangxinghan/meta-faster-r-cnn}}, TFA\footnote{\url{https://github.com/ucbdrive/few-shot-object-detection}}, FSCE\footnote{\url{https://github.com/megvii-research/FSCE}}, DeFRCN\footnote{\url{https://github.com/er-muyue/DeFRCN}}. FRCN-ft is a Faster R-CNN \cite{ren2016faster} detector which is simply trained on the base dataset, then fine-tuned on the $K$-shot target instances. The teacher is used for the inference and evaluation of test images. 
\subsection{Main Results} 

As shown in Table \ref{tab:2}, our proposed approach outperforms existing FSOD approaches in all settings. Overall, our approach produces an average 2.0\% improvement over the second-best approach on the three datasets. We further observe that all approaches obtain performance gains without freezing parameters. This confirm that freezing some parameters \cite{guo2020broader,tseng2020cross,islam2021dynamic} can not alleviate the overfitting problem in the CD-FSOD. Moreover, these approaches still do not show satisfactory performance. The meta-learning approaches still fail to outperform naive fine-tuned models in all settings. This suggests that meta-learning approaches use supervision for pre-training and cannot mimic distant domain datasets, which leads them to overfit the source data and generalize poorly to distant target domains. The fine-tuning approaches DeFRCN and FSCE have only a slight performance improvement over FRCN-ft. This suggests that these approaches tailored for FSOD do not work with CD-FSOD. There is a desirable need to develop approaches that work under both FSOD and CD-FSOD.
 \begin{table*}[!htb] 
\begin{center}
 \resizebox{\linewidth}{!}{
  \begin{tabular}{c|ccc|ccc|ccc|c}
    \toprule
          \multirow{2}{*}{Method/Shot}& \multicolumn{3}{c|}{ArTaxOr} & \multicolumn{3}{c|}{UODD}& \multicolumn{3}{c|}{DIOR}& \multirow{2}{*}{Avg}\\
     &1&  5 & 10&1 &5 & 10&1 &5 & 10\\
       \midrule
       A-RPN  \cite{wang2020frustratingly} &  2.5$_{{\color{red} - 1.1} }$ & 8.1$_{{\color{red} - 3.1}}$  & 13.9$_{{\color{red} - 4.2}}$ &  3.3$_{{\color{red} - 1.0}}$ & 8.4$_{{\color{red} - 2.3}}$  & 10.8 $_{{\color{red} - 1.6}}$ &  7.5$_{{\color{red} - 1.2}}$ & 17.1$_{{\color{red} - 2.7}}$  & 20.3$_{{\color{red} - 2.4}}$ & 10.2$_{{\color{red} - 2.1}}$\\
    Meta-RCNN \cite{han2022meta}  & 2.8$_{{\color{red} - 0.9}}$  & 8.5$_{{\color{red} - 2.4}}$ &  14.0$_{{\color{red} - 3.7}}$ & 3.6$_{{\color{red} - 0.8}}$  & 8.8$_{{\color{red} - 2.1}}$ & 11.2$_{{\color{red} - 1.3}}$& 7.8 $_{{\color{red} - 2.3}}$& 17.7$_{{\color{red} - 2.5}}$&  20.6$_{{\color{red} - 1.8}}$ & 10.6$_{{\color{red} - 1.9}}$ \\ 
    H-GCN\cite{han2021query}& 2.6$_{{\color{red} - 0.6}}$  & 8.2$_{{\color{red} - 1.9}}$ & 14.2$_{{\color{red} - 3.3}}$& 3.8$_{{\color{red} - 0.7}}$ & 7.7$_{{\color{red} - 1.5}}$& 11.0 $_{{\color{red} - 1.6}}$& 7.9$_{{\color{red} - 1.9}}$ & 18.0$_{{\color{red} - 2.6}}$& 20.9$_{{\color{red} - 2.2}}$ &10.5$_{{\color{red} - 1.8}}$\\
    TFA w/cos \cite{fan2020few}& 3.1$_{{\color{red} - 2.3}}$ & 8.8$_{{\color{red} - 5.0}}$  & 14.8$_{{\color{red} - 7.7}}$& 4.4$_{{\color{red} - 1.7}}$ &  8.7 $_{{\color{red} - 2.2}}$& 11.8$_{{\color{red} - 4.6}}$ & 8.0$_{{\color{red} - 4.1}}$ &  18.1$_{{\color{red} - 7.8}}$&  20.5 $_{{\color{red} - 7.1}}$ &10.9 $_{{\color{red} - 4.7}}$\\
    FSCE  \cite{DBLP:conf/cvpr/SunLCYZ21}  & 3.7$_{{\color{red} - 1.9}}$ & 10.2$_{{\color{red} - 4.3}}$ & 15.9 $_{{\color{red} - 5.1}}$ & 3.9$_{{\color{red} - 1.1}}$& 9.6 $_{{\color{red} - 2.9}}$& 12.0$_{{\color{red} - 3.6}}$& 8.6$_{{\color{red} - 3.0}}$& 18.7$_{{\color{red} - 3.8}}$& 21.9$_{{\color{red} - 3.6}}$&11.6$_{{\color{red} - 3.2}}$\\ 
    DeFRCN  \cite{qiao2021defrcn}  & 3.6$_{{\color{red} - 0.7}}$ &9.9$_{{\color{red} - 1.1}}$  &  15.5$_{{\color{red} - 1.0}}$  & 4.5$_{{\color{red} - 0.8}}$ & 9.9$_{{\color{red} - 1.0}}$ & 12.1$_{{\color{red} - 1.4}}$& 9.3$_{{\color{red} - 1.3}}$ & 18.9$_{{\color{red} - 1.2}}$& 22.9$_{{\color{red} - 2.2}}$ &11.8$_{{\color{red} - 1.2}}$\\ 
    FRCN-ft   & 3.4 & 9.3 & 15.2 & 4.1& 9.2 & 12.3& 8.4& 18.3& 21.2& 11.2\\ 
      Ours & \textbf{5.1}& \textbf{12.5}& \textbf{18.1}& \textbf{5.9}& \textbf{12.2}& \textbf{14.5} & \textbf{10.5} &\textbf{19.1} & \textbf{26.5} & \textbf{13.8}\\
    \bottomrule
  \end{tabular}}
  \end{center}
    \vspace{-15pt}
    \caption{The performance (mAP) on the CD-FSOD benchmark. The best results are in bold. Red numbers indicate performance degradation due to frozen parameters.}
      \label{tab:2}
\end{table*}

\subsection{Ablation Studies} 
In this section, we show the ablation experiments on the DIOR dataset.

\begin{table}[!htb]
\centering
 \resizebox{\linewidth}{!}{
 \begin{tabular}{cc|cc|cc|cc}
    \toprule
 \multirow{2}{*}{ EMA}& \multirow{2}{*}{distillation} & \multicolumn{2}{c|}{1}&  \multicolumn{2}{c|}{5} & \multicolumn{2}{c}{10}\\
     \cmidrule{3-8}

      & &  S & T &  S & T &  S & T  \\
      \midrule
      \checkmark& & 8.4 & 9.2 &18.3& 18.7 & 21.2& 24.8\\
      & \checkmark & 8.6 & 8.9 & 18.4 & 18.5 &22.3 &23.1\\ 
      \checkmark&\checkmark &9.5& 10.5 & 18.8 &19.1 & 23.6&26.5\\ 
    \bottomrule
  \end{tabular}}
    \vspace{-5pt}
  \caption{The effect of EMA and the distillation. ``S'' and ``T'' represent the student and the teacher respectively.}
  \label{tab:3}
  \vspace{-10pt}
\end{table}

\begin{figure}[!htp]
\flushleft

    \subfloat[\centering ]{
\begin{tikzpicture}[scale=0.33] 
\pgfplotsset{compat=1.17}
\begin{axis}[
    xtick={20,40,60,80,100},
    xticklabels={0.5,0.7,0.9,0.999,0.9999},
    xlabel= $\alpha$,
    ytick={0,5,10,15,20,25,30},
    ymajorgrids=true,
     ylabel=mAP,
    grid style=dashed,
     legend style={at={(0.8,0.5)},anchor=north} %
    ]

\addplot[smooth,mark=*,blue] plot coordinates { 
    (20,20.9)
    (40,23.9 )
    (60,24.9)
    (80,26.4)
    (100,25.7)

};

\addlegendentry{10-shot}

\addplot[smooth,mark=triangle*,cyan] plot coordinates {
    (20,14.8)
    (40,16.1)
    (60,17.2)
    (80,19.1)
    (100,18.4)

};
\addlegendentry{5-shot}

\addplot[smooth,mark=square*,red] plot coordinates {
    (20,8.4)
    (40,9.2)
    (60,9.6)
    (80,10.5)
    (100,9.3)
};
\addlegendentry{1-shot}
\end{axis}
\end{tikzpicture}

}
    \subfloat[\centering ]{
  \begin{tikzpicture}[scale=0.33]
  \pgfplotsset{compat=1.17}
\begin{axis}[
    xtick={20,40,60,80},
    xticklabels={0.6,0.7,0.8,0.9},
    xlabel= $\delta$, 
    ytick={0,5,10,15,20,25,30},
    ymajorgrids=true,
     ylabel=mAP,
    grid style=dashed,
     legend style={at={(0.8,0.42)},anchor=north} %
    ]
\addplot[smooth,mark=*,blue] plot coordinates { 
    (20,24.9)
    (40,26.5)
    (60,23.2)
    (80,22.9)
};

\addlegendentry{10-shot}

\addplot[smooth,mark=triangle*,cyan] plot coordinates {
    (20,16.9)
    (40,19.1)
    (60,17.4)
    (80,17.1)
};
\addlegendentry{5-shot}

\addplot[smooth,mark=square*,red] plot coordinates {
    (20,9.0)
    (40,10.5)
    (60,9.3)
    (80,9.5)

};
\addlegendentry{1-shot}

\end{axis}
\end{tikzpicture}
    }
    \subfloat[\centering ]{
\begin{tikzpicture}[scale=0.33] 
\pgfplotsset{compat=1.17}
\begin{axis}[
    xtick={20,40,60,80,100,120,140},
    xticklabels={1.0,2.0,3.0,4.0,5.0,6.0,7.0},
    xlabel= $\lambda$, 
    ytick={0,5,10,15,20,25,30},
    ymajorgrids=true,
     ylabel=mAP,
    grid style=dashed,
     legend style={at={(0.8,0.46)},anchor=north} %
    ]
\addplot[smooth,mark=*,blue] plot coordinates { 
    (20,23.1)
    (40,24.5)
    (60,24.9)
    (80,26.5)
    (100,25.4)
    (120,25.1)
    (140,24.6)

};
\addlegendentry{10-shot}

\addplot[smooth,mark=triangle*,cyan] plot coordinates {
    (20,15.5)
    (40,17.4)
    (60,19.1)
    (80,18.5)
    (100,18.0)
    (120,18.1)
    (140,17.7)
};
\addlegendentry{5-shot}

\addplot[smooth,mark=square*,red] plot coordinates {
    (20,8.1)
    (40,9.4)
    (60,10.1)
    (80,10.5)
    (100,9.9)
    (120,9.5)
    (140,9.3)

};
\addlegendentry{1-shot}
\end{axis}
\end{tikzpicture}}
  \vspace{-8pt}
    \caption{ Ablation studies for (a) EMA rate $\alpha$, (b) pseudo-labeling threshold $\delta $, and (d) distillation loss weights $\lambda$.}
    \label{fig:3}
\end{figure}
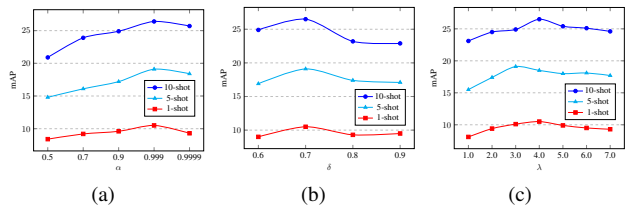

\noindent \textbf{EMA and Distillation. } As shown in Table \ref{tab:3}, the EMA and distillation both improve the performance but EMA achieves better performance than distillation. The combination of distillation and EMA, leads to even better performance. We also observe that the performance of the students increases with the performance of the teacher. This means that our proposed approach enables students and teachers to progress together in a mutually beneficial manner. Extensive research has shown that there is an overfitting problem in FSOD. We argue that EMA and distillation can effectively alleviate the problem in FSOD. With EMA, the teacher can be seen as an average model of the student over different steps, so it is more stable and robust. And the distillation loss can be seen as a regularization, which can improve the generalization of the student.

\noindent \textbf {EMA rate.} We evaluate the model using various EMA rates $\alpha$ from 0.5 to 0.9999, and present the mAP results in Figure \ref{fig:3} (a). When the EMA ratio is small (e.g., $\alpha$ = 0.5), the student contributes more to the teacher model in each iteration, which leads to an unstable teacher model with a lower mAP.  This situation can be stabilized and improved as the EMA ratio $\alpha$ increases. It performs the best mAP when the EMA ratio $\alpha$ reaches 0.999. However, if the EMA rate $\alpha$ keeps increasing, the teacher model performance will degrade because the teacher model mainly derives the next model weights from the previous teacher model weights.
 
\noindent \textbf {Pseudo-labeling thresholding.} Pseudo-labeling thresholding plays an important role in the distillation loss, as it can filter the low-confidence predicted bounding boxes. As shown in Figure \ref{fig:3} (b), if the threshold is too low (e.g. $\alpha$ = 0.6), the mAP of the model is low because the model predicts more unreliable bounding boxes. On the other hand, the performance of a model using an excessively high threshold (e.g., $\alpha$  = 0.9) degrades as it cannot predict a sufficient number of bounding boxes in its generated pseudo-labels.

\noindent \textbf {Distillation loss weight.} To examine the effect of the distillation loss weight, we vary the distillation loss weight $\lambda$ from 1.0 to 8.0. As shown in Figure \ref{fig:3} (c), with a lower distillation loss weight $\lambda$  = 1.0, the model has a lower performance. On the other hand, we observe that the model performs the best with the loss weight $\lambda$ = 4.0 (1-shot and 10-shot) or 3.0 (5-shot).

\begin{figure}[!htb]
\centering
 \includegraphics[width=  0.95\linewidth]{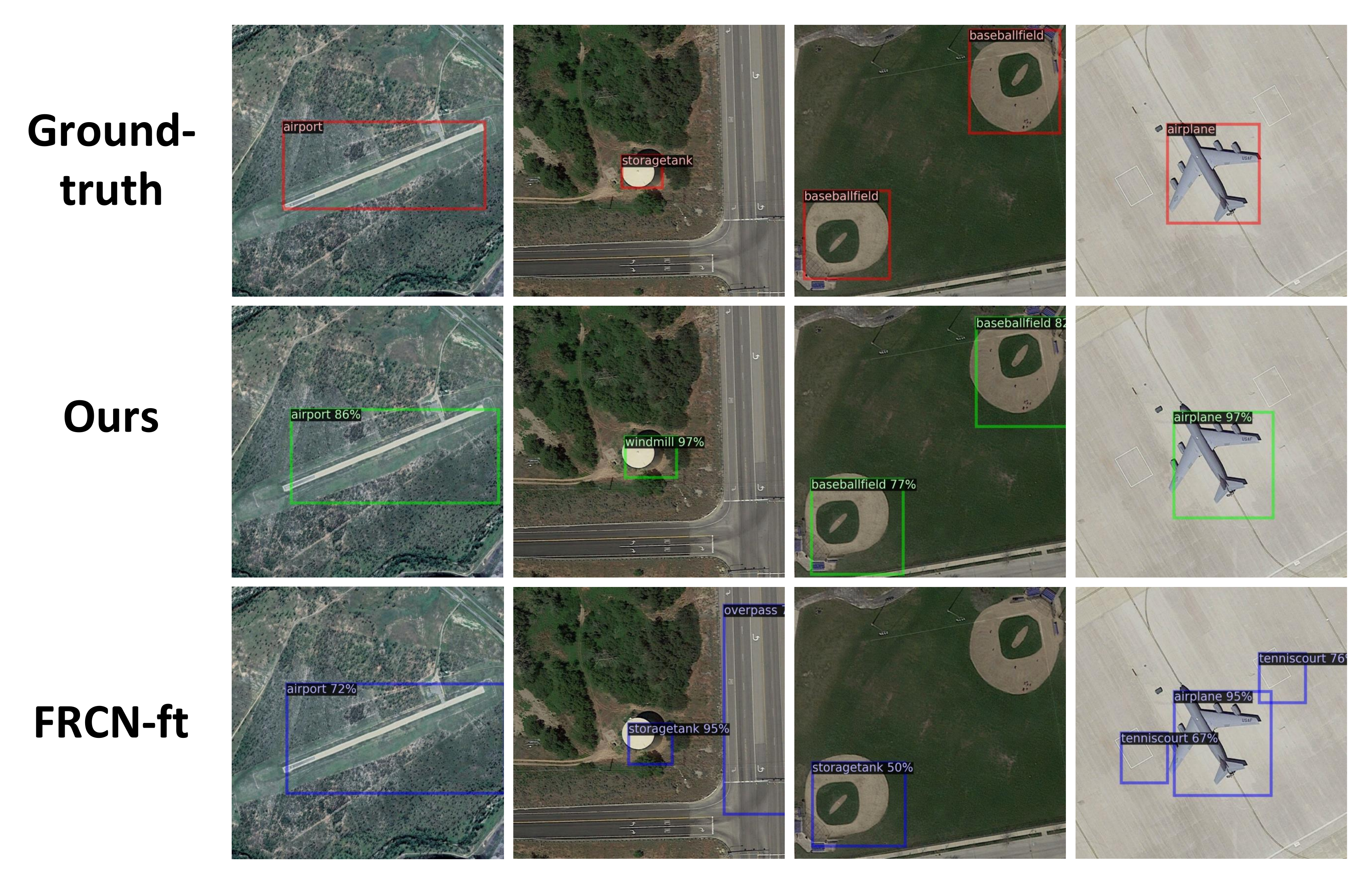}
 \vspace{-8pt}
\caption{Detection examples based on 10-shot setting. FRCN-ft leads to three types of false detections: missed detection of certain objects (the third example in the third row), incorrect detection of background (the second example in the third row) and inaccurate localization (the first example in the third row). For these examples, our approach reduces the occurrence of these errors.}
\label{fig:4}
\vspace{-8pt}
\end{figure}

\section{Conclusion}  
In this paper, we formally introduce the study of the cross-domain few-shot object detection (CD-FSOD) benchmark, which covers several target domains with varying similarities to natural images. On the proposed benchmarks, we evaluate existing FSOD approaches and analyze the reasons for their failure. Then, we introduce a strong baseline that achieves state-of-the-art performance on the proposed benchmark. In the future, we will work on developing novel approaches for both FSOD and CD-FSOD.

\noindent \textbf{Acknowledgement. }The authors wish to acknowledge CSC IT Center for Science, Finland, for computational resources.

\clearpage


\bibliographystyle{IEEEbib}
\bibliography{strings,refs}

\end{document}